# ChrEn: Cherokee-English Machine Translation for Endangered Language Revitalization

**Shiyue Zhang**  **Benjamin Frey**  **Mohit Bansal**
UNC Chapel Hill
{shiyue, mbansal}@cs.unc.edu; benfrey@email.unc.edu

## Abstract

Cherokee is a highly endangered Native American language spoken by the Cherokee people. The Cherokee culture is deeply embedded in its language. However, there are approximately only 2,000 fluent first language Cherokee speakers remaining in the world, and the number is declining every year. To help save this endangered language, we introduce *ChrEn*, a Cherokee-English parallel dataset, to facilitate machine translation research between Cherokee and English. Compared to some popular machine translation language pairs, ChrEn is extremely low-resource, only containing 14k sentence pairs in total. We split our parallel data in ways that facilitate both in-domain and out-of-domain evaluation. We also collect 5k Cherokee monolingual data to enable semi-supervised learning. Besides these datasets, we propose several Cherokee-English and English-Cherokee machine translation systems. We compare SMT (phrase-based) versus NMT (RNN-based and Transformer-based) systems; supervised versus semi-supervised (via language model, back-translation, and BERT/Multilingual-BERT) methods; as well as transfer learning versus multilingual joint training with 4 other languages. Our best results are 15.8/12.7 BLEU for in-domain and 6.5/5.0 BLEU for out-of-domain Chr-En/En-Chr translations, respectively, and we hope that our dataset and systems will encourage future work by the community for Cherokee language revitalization.[1]

| Src. | iɬ RGə DЛə ɴɥY, ӨᴔYӕ DB RGə FT hFRΘ FrY. |
|---|---|
| Ref. | They are not of the world, even as I am not of the world. |
| SMT | It was not the things upon the earth, even as I am not of the world. |
| NMT | I am not the world, even as I am not of the world. |

Table 1: An example from the development set of *ChrEn*. NMT denotes our RNN-NMT model.

## 1 Introduction

The Cherokee people are one of the indigenous peoples of the United States. Before the 1600s, they lived in what is now the southeastern United States (Peake Raymond, 2008). Today, there are three federally recognized nations of Cherokee people: the Eastern Band of Cherokee Indians (EBCI), the United Keetoowah Band of Cherokee Indians (UKB), and the Cherokee Nation (CN). The Cherokee language, the language spoken by the Cherokee people, contributed to the survival of the Cherokee people and was historically the basic medium of transmission of arts, literature, traditions, and values (Nation, 2001; Peake Raymond, 2008). However, according to the Tri-Council Res. No. 02-2019, there are only 2,000 fluent first language Cherokee speakers left, and each Cherokee tribe is losing fluent speakers at faster rates than new speakers are developed. UNESCO has identified the dialect of Cherokee in Oklahoma is "definitely endangered", and the one in North Carolina is "severely endangered". Language loss is the loss of culture. CN started a 10-year language revitalization plan (Nation, 2001) in 2008, and the Tri-Council of Cherokee tribes declared a state of emergency in 2019 to save this dying language.

To revitalize Cherokee, language immersion programs are provided in elementary schools, and second language programs are offered in universities. However, students have difficulty finding exposure to this language beyond school hours (Albee, 2017). This motivates us to build up English (En) to Cherokee (Chr) machine translation systems so that we could automatically translate or aid human translators to translate English materials to Cherokee. Chr-to-En is also highly meaningful in helping spread Cherokee history and culture. Therefore, in this paper, we contribute our effort to Cherokee revitalization by constructing a clean **Cher**okee-**En**glish parallel dataset, **ChrEn**, which results in 14,151 pairs of sentences with around 313K English tokens and 206K Cherokee tokens. We also collect 5,210 Cherokee monolingual sentences with 93K Cherokee tokens. Both datasets are derived from bilingual or monolingual materials that are translated or written by first-language Cherokee speakers, then we manually aligned and cleaned the raw data.[2] Our datasets contain texts of two Cherokee dialects (Oklahoma and North Carolina), and diverse text types (e.g., sacred text, news). To facilitate the development of machine translation systems, we split our parallel data into five subsets: Train/Dev/Test/Out-dev/Out-test, in which Dev/Test and Out-dev/Out-test are for in-domain and out-of-domain evaluation respectively. See an example from *ChrEn* in Table 1 and the detailed dataset description in Section 3.

---
[1] Our data, code, and demo will be publicly available at https://github.com/ZhangShiyue/ChrEn.

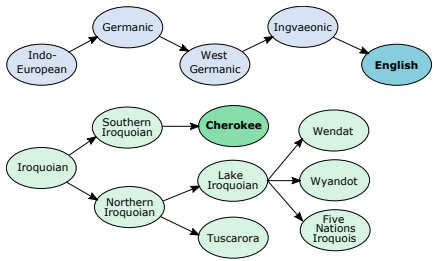

Figure 1: Language family trees.

The translation between Cherokee and English is not easy because the two languages are genealogically disparate. As shown in Figure 1, Cherokee is the sole member of the southern branch of the Iroquoian language family and is unintelligible to other Iroquoian languages, while English is from the West Germanic branch of the Indo-European language family. Cherokee uses a unique 85-character syllabary invented by Sequoyah in the early 1820s, which is highly different from English's alphabetic writing system. Cherokee is a polysynthetic language, meaning that words are composed of many morphemes that each have independent meanings. A single Cherokee word can express the meaning of several English words, e.g., ᎣᏧᏂᏁᎩᏏ (widatsinegisi), or *I am going off at a distance to get a liquid object*. Since the semantics are often conveyed by the rich morphology, the word orders of Cherokee sentences are variable. There is no "basic word order" in Cherokee, and most word orders are possible (Montgomery-Anderson, 2008), while English generally follows the Subject-Verb-Object (SVO) word order. Plus, verbs comprise 75% of Cherokee, which is only 25% for English (Feeling, 1975, 1994).

Hence, to develop translation systems for this low-resource and distant language pair, we investigate various machine translation paradigms and propose phrase-based (Koehn et al., 2003) Statistical Machine Translation (SMT) and RNN-based (Luong et al., 2015) or Transformer-based (Vaswani et al., 2017) Neural Machine Translation (NMT) systems for both Chr-En and En-Chr translations, as important starting points for future works. We apply three semi-supervised methods: using additional monolingual data to train the language model for SMT (Koehn and Knowles, 2017); incorporating BERT (or Multilingual-BERT) (Devlin et al., 2019) representations for NMT (Zhu et al., 2020), where we introduce four different ways to use BERT; and the back-translation method for both SMT and NMT (Bertoldi and Federico, 2009; Lambert et al., 2011; Sennrich et al., 2016b). Moreover, we explore the use of existing X-En parallel datasets of 4 other languages (X = Czech/German/Russian/Chinese) to improve Chr-En/En-Chr performance via transfer learning (Kocmi and Bojar, 2018) or multilingual joint training (Johnson et al., 2017).

Empirically, NMT is better than SMT for in-domain evaluation, while SMT is significantly better under the out-of-domain condition. RNN-NMT consistently performs better than Transformer-NMT. Semi-supervised learning improves supervised baselines in some cases (e.g., back-translation improves out-of-domain Chr-En NMT by 0.9 BLEU). Even though Cherokee is not related to any of the 4 languages (Czech/German/Russian/Chinese) in terms of their language family trees, surprisingly, we find that both transfer learning and multilingual joint training can improve Chr-En/En-Chr performance in most cases. Especially, transferring from Chinese-English achieves the best in-domain Chr-En performance, and joint learning with English-German obtains the best in-domain En-Chr performance. The best results are 15.8/12.7 BLEU for in-domain Chr-En/En-Chr translations; and 6.5/5.0 BLEU for out-of-domain Chr-En/En-Chr translations. Finally, we conduct a 50-example human (expert) evaluation; however, the human judgment does not correlate with BLEU for the En-Chr translation, indi-cating that BLEU is possibly not very suitable for Cherokee evaluation. Overall, we hope that our datasets and strong initial baselines will encourage future works to contribute to the revitalization of this endangered language.

---
[2] Our co-author, Prof. Benjamin Frey, is a proficient second-language Cherokee speaker and a citizen of the Eastern Band of Cherokee Indians.

## 2 Related Works

**Cherokee Language Revitalization.** In 2008, the Cherokee Nation launched the 10-year language preservation plan (Nation, 2001), which aims to have 80% or more of the Cherokee people be fluent in this language in 50 years. After that, a lot of revitalization works were proposed. Cherokee Nation and the EBCI have established language immersion programs and k-12 language curricula. Several universities, including the University of Oklahoma, Stanford University, etc., have begun offering Cherokee as a second language. However, given Cherokee has been rated at the highest level of learning difficulty (Peake Raymond, 2008), it is hard to be mastered without frequent language exposure. As mentioned by Crystal (2014), an endangered language will progress if its speakers can make use of electronic technology. Currently, the language is included among existing Unicode-compatible fonts, is supported by Gmail, and has a Wikipedia page. To revitalize Cherokee, a few Cherokee pedagogical books have been published (Holmes and Smith, 1976; Joyner, 2014), as well as several online learning platforms.[3] Feeling (2018) provided detailed English translations and linguistic analysis of a number of Cherokee stories. A Digital Archive for American Indian Languages Preservation and Perseverance (DAILP) has been developed for transcribing, translating, and contextualizing historical Cherokee language documents (Bourns, 2019; Cushman, 2019).[4] However, the translation between Cherokee and English still can only be done by human translators. Given that only 2,000 fluent first-language speakers are left, and the majority of them are elders, it is important and urgent to have a machine translation system that could assist them with translation. Therefore, we introduce a clean Cherokee-English parallel dataset to facilitate machine translation development and propose multiple translation systems as starting points of future works. We hope our work could attract more attention from the NLP community in helping to save and revitalize this endangered language. An initial version of our data and its implications was introduced in (Frey, 2020). Note that we are not the first to propose a Cherokee-English parallel dataset. There is Chr-En parallel data available on OPUS (Tiedemann, 2012).[5] The main difference is that our parallel data contains 99% of their data and has 6K more examples from diverse domains.

**Low-Resource Machine Translation.** Even though machine translation has been studied for several decades, the majority of the initial research effort was on high-resource translation pairs, e.g., French-English, that have large-scale parallel datasets available. However, most of the language pairs in the world lack large-scale parallel data. In the last five years, there is an increasing research interest in these low-resource translation settings. The DARPA's LORELEI language packs contain the monolingual and parallel texts of three dozen languages that are considered as low-resource (Strassel and Tracey, 2016). Riza et al. (2016) proposed several low-resource Asian language pairs. Lakew et al. (2020) and Duh et al. (2020) proposed benchmarks for five and two low-resource African languages, respectively. Guzmán et al. (2019) introduced two low-resource translation evaluation benchmarks: Nepali–English and Sinhala–English. Besides, most low-resource languages rely on the existing parallel translations of the Bible (Christodouloupoulos and Steedman, 2015). Because not many low-resource parallel datasets were publicly available, some low-resource machine translation research was done by sub-sampling high-resource language pairs (Johnson et al., 2017; Lample et al., 2018), but it may downplay the fact that low-resource translation pairs are usually distant languages. Our *ChrEn* dataset can not only be another open resource of low-resource MT research but also challenge MT methods with an extremely morphology rich language and a distant language pair. Two methods have been largely explored by existing works to improve low-resource MT. One is semi-supervised learning to use monolingual data (Gulcehre et al., 2015; Sennrich et al., 2016b). The other is cross-lingual transfer learning or multilingual joint learning (Kocmi and Bojar, 2018; Johnson et al., 2017). We explore both of them to improve Chr-En/En-Chr translations.

---

[3] `mangolanguages.com/available-languages/learn-cherokee/`, `yourgrandmotherscherokee.com`
[4] `https://dailp.northeastern.edu/`
[5] `http://opus.nlpl.eu/`

| Statistics | Parallel | | | | | | Monolingual |
| --- | --- | --- | --- | --- | --- | --- | --- |
| | Train | Dev | Test | Out-dev | Out-test | Total | Total |
| Sentences (or Sentence pairs) | 11,639 | 1,000 | 1,000 | 256 | 256 | 14,151 | 5,210 |
| English tokens | 257,460 | 21,686 | 22,154 | 5,867 | 6,020 | 313,187 | - |
| Unique English tokens | 11,606 | 3,322 | 3,322 | 1,605 | 1,665 | 13,621 | - |
| % Unseen unique English tokens | - | 13.3 | 13.2 | 42.1 | 43.3 | - | - |
| Average English sentence length | 22.1 | 21.7 | 22.2 | 22.9 | 23.5 | 22.1 | - |
| Cherokee tokens | 168,389 | 14,367 | 14,373 | 4,324 | 4,370 | 205,823 | 92,897 |
| Unique Cherokee tokens | 32,419 | 5,182 | 5,244 | 1,857 | 1,881 | 38,494 | 19,597 |
| % Unseen unique Cherokee tokens | - | 37.7 | 37.3 | 67.5 | 68.0 | - | 73.7 |
| Average Cherokee sentence length | 14.5 | 14.4 | 14.3 | 16.9 | 17.1 | 14.5 | 17.8 |

Table 2: The key statistics of our parallel and monolingual data. Note that "% Unseen unique English tokens" is in terms of the Train split, for example, 13.3% of unique English tokens in Dev are unseen in Train.

## 3 Data Description

It is not easy to collect substantial data for endangered Cherokee. We obtain our data from bilingual or monolingual books and newspaper articles that are translated or written by first-language Cherokee speakers. In the following, we will introduce the data sources and the cleaning procedure and give detailed descriptions of our data statistics.

### 3.1 Parallel Data

Fifty-six percent of our parallel data is derived from the *Cherokee New Testament*. Other texts are novels, children's books, newspaper articles, etc. These texts vary widely in dates of publication, the oldest being dated to 1860. Additionally, our data encompasses both existing dialects of Cherokee: the Overhill dialect, mostly spoken in Oklahoma (OK), and the Middle dialect, mostly used in North Carolina (NC). These two dialects are mainly phonologically different and only have a few lexical differences (Uchihara, 2016). In this work, we do not explicitly distinguish them during translation. The left pie chart of Figure 2 shows the parallel data distributions over text types and dialects, and the complete information is in Table 14 of Appendix A.1. Many of these texts were translations of English materials, which means that the Cherokee structures may not be 100% natural in terms of what a speaker might spontaneously produce. But each text was translated by people who speak Cherokee as the first language, which means there is a high probability of grammaticality. These data were originally available in PDF version. We apply the Optical Character Recognition (OCR) via Tesseract OCR engine[6] to extract the Cherokee and English text. Then our co-author, a proficient second-language speaker of Cherokee, manually aligned the sentences and fixed the errors introduced by OCR. This process is time-consuming and took several months.

[6]https://github.com/tesseract-ocr/

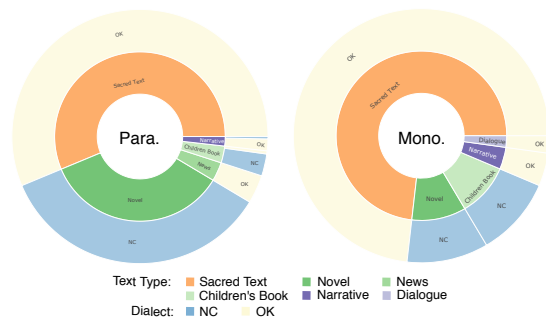

Figure 2: The distributions of our parallel (Para.) and monolingual (Mono.) data over text types and dialects.

The resulting dataset consists of 14,151 sentence pairs. After tokenization,[7] there are around 313K English tokens and 206K Cherokee tokens in total with 14K unique English tokens and 38K unique Cherokee tokens. Notably, the Cherokee vocabulary is much larger than English because of its morphological complexity. This casts a big challenge to machine translation systems because a lot of Cherokee tokens are infrequent. To facilitate machine translation system development, we split this data into training, development, and testing sets. As our data stems from limited sources, we find that if we randomly split the data, some phrases/sub-sentences are repeated in training and evaluation sets, so the trained models will overfit to these frequent patterns. Considering that low-resource translation is usually accompanied by out-of-domain generalization in real-world applications, we provide two groups of develop-ment/testing sets. We separate all the sentence pairs from newspaper articles, 512 pairs in total, and randomly split them in half as out-of-domain development and testing sets, denoted by **Out-dev** and **Out-test**. The remaining sentence pairs are randomly split into in-domain **Train**, **Dev**, and **Test**. About 13.3% of unique English tokens and 37.7% of unique Cherokee tokens in Dev have not appeared in Train, while the percentages are 42.1% and 67.5% for Out-dev, which shows the difficulty of the out-of-domain generalization. Table 2 contains more detailed statistics; notably, the average sentence length of Cherokee is much shorter than English, which demonstrates that the semantics are morphologically conveyed in Cherokee.

[7]We tokenize both English and Cherokee by Moses tokenizer (Koehn et al., 2007). For Cherokee, it is equivalent to tokenize by whitespace and punctuation, confirmed to be good enough by our Cherokee-speaker coauthor.

Note that Cherokee-English parallel data is also available on OPUS (Tiedemann, 2012), which has 7.9K unique sentence pairs, 99% of which are the *Cherokee New Testament* that are also included in our parallel data, i.e., our data is bigger and has 6K more sentence pairs that are not sacred texts (novels, news, etc.). The detailed comparison will be discussed in A.2.

### 3.2 Monolingual Data

In addition to the parallel data, we also collect a small amount of Cherokee monolingual data, 5,210 sentences in total. This data is also mostly derived from Cherokee monolingual books.[8] As depicted by the right pie chart in Figure 2, the majority of monolingual data are also sacred text, which is *Cherokee Old Testament*, and it also contains two-dialect Cherokee texts. Complete information is in Table 15 of Appendix A.1. Similarly, we applied OCR to extract these texts. However, we only manually corrected the major errors introduced by OCR. Thus our monolingual data is noisy and contains some lexical errors. As shown in Table 2, there are around 93K Cherokee tokens in total with 20K unique Cherokee tokens. This monolingual data has a very small overlap with the parallel data; about 72% of the unique Cherokee tokens are unseen in the whole parallel data. Note that most of our monolingual data have English translations, i.e., it could be converted to parallel data. But it requires more effort from Cherokee speakers and will be part of our future work. For now, we show how to effectively use this monolingual data for semi-supervised gains.

---

[8]We considered parsing Cherokee Wikipedia. But, according to our coauthor, who is a Cherokee speaker, its content is mostly low-quality.

## 4 Models

In this section, we will introduce our Cherokee-English and English-Cherokee translation systems. Adopting best practices from low-resource machine translation works, we propose both Statistical Machine Translation (SMT) and Neural Machine Translation (NMT) systems, and for NMT, we test both RNN-based and Transformer-based models. We apply three semi-supervised methods: training language model with additional monolingual data for SMT (Koehn and Knowles, 2017), incorporating BERT or Multilingual-BERT representations into NMT (Zhu et al., 2020), and back-translation for both SMT and NMT (Bertoldi and Federico, 2009; Sennrich et al., 2016b). Further, we explore transfer learning (Kocmi and Bojar, 2018) from and multilingual joint training (Johnson et al., 2017) with 4 other languages (Czech/German/Russian/Chinese) for NMT.

### 4.1 SMT

**Supervised SMT.** SMT was the mainstream of machine translation research before neural models came out. Even if NMT has achieved state-of-the-art performance on many translation tasks, SMT is still very competitive under low-resource and out-of-domain conditions (Koehn and Knowles, 2017). Phrase-based SMT is a dominant paradigm of SMT (Koehn et al., 2003). It first learns a phrase table from the parallel data that translates source phrases to target. Then, a reordering model learns to reorder the translated phrases. During decoding, a scoring model scores candidate translations by combining the weights from translation, reordering, and language models, and it is tuned by maximizing the translation performance on the development set. A simple illustration of SMT is shown in Figure 3. Note that, as Cherokee and English have different word orders (English follows SVO; Cherokee has variable word orders), one Cherokee phrase could be translated into two English words that are far apart in the sentence. This increases the difficulty of SMT that relies on phrase correspondence and is not good at distant word reordering (Zhang et al., 2017). We implement our SMT systems by Moses (Koehn et al., 2007).

**Semi-Supervised SMT.** Previous works have shown that SMT can be improved by two semi-supervised methods: (1) A big language model (Koehn and Knowles, 2017), i.e., a language model trained with big target-side monolingual data; (2) Synthesizing bilingual data by back-translating monolingual data (Bertoldi and Federico, 2009; Lambert et al., 2011). Using our Cherokee monolingual data and the publicly available English monolingual data, we test these two methods. For the first method, we use both parallel and monolingual data to train the language model; for the second method, we back-translate target-language monolingual data into the source language and then combine them with the training set to retrain a source-target SMT model.

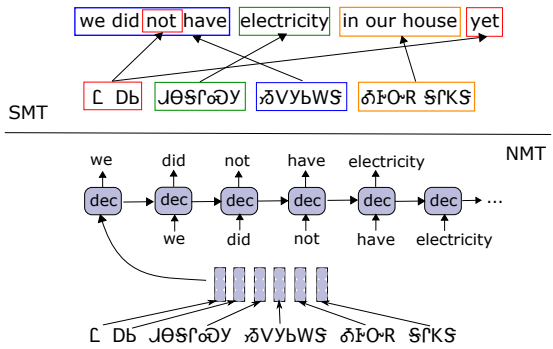

Figure 3: A simple illustration of SMT and NMT.

### 4.2 NMT

**Supervised NMT.** NMT has mostly dominated recent machine translation research. Especially when a large amount of parallel data is available, NMT surpasses SMT by a large margin; moreover, NMT is good at generating fluent translations because of its auto-regressive generation nature. Koehn and Knowles (2017) pointed out the poor performance of NMT under low-resource and out-of-domain conditions; however, recent work from Sennrich and Zhang (2019) showed that low-resource NMT can be better than SMT by using proper training techniques and hyper-parameters. NMT models usually follow encoder-decoder architecture. The encoder encodes the source sentence into hidden representations, then the decoder generates the target sentence word by word by "reading" these representations, as shown in Figure 3. We investigate two paradigms of NMT implementations: RNN-based model (Bahdanau et al., 2015) and Transformer-based model (Vaswani et al., 2017). We implement both of them via OpenNMT (Klein et al., 2017). For RNN-NMT, we follow the global attentional model with general attention proposed by Luong et al. (2015). For Transformer-NMT, we mainly follow the architecture proposed by Vaswani et al. (2017) except applying layer normalization before the self-attention and FFN blocks instead of after, which is more robust (Baevski and Auli, 2019).

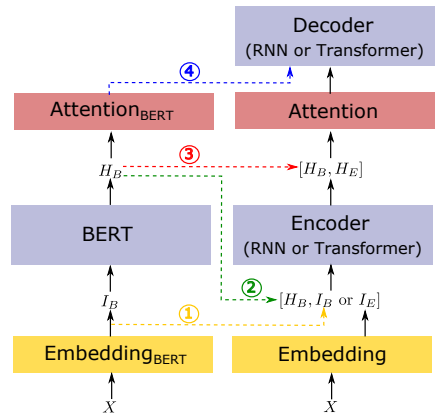

Figure 4: The four different ways we proposed to incorporate BERT representations into NMT models.

**Semi-Supervised NMT.** NMT models can often be improved when more training data is available; therefore, a lot of works have studied semi-supervised approaches that utilize monolingual data to improve translation performance. Similar to SMT, we mainly investigate two semi-supervised methods. The first is to leverage pre-trained language models. Early works proposed shallow or deep fusion methods to rerank NMT outputs or add the language model's hidden states to NMT decoder (Jean et al., 2015; Gulcehre et al., 2015). Recently, the large-scale pre-trained language model, BERT (Devlin et al., 2019), has achieved impressive success in many NLP tasks. Zhu et al. (2020) showed that incorporating the contextualized BERT representations can significantly improve translation performances. Following but different from this work, we explore four different ways to incorporate BERT representations into NMT models for English-Cherokee translation only.[9] As depicted in Figure 4, we apply BERT representations by: ① Initializing NMT models' word embedding matrix with BERT's pre-trained word embedding matrix $I_B$; ② Concatenating NMT encoder's input $I_E$ with BERT's output $H_B$; ③ Concatenating NMT encoder's output $H_E$ with BERT's output $H_B$; ④ Using another attention to leverage BERT's output $H_B$ into decoder. Note that ③ and ④ will not be applied simultaneously, and all the combination of these four methods are treated as hyper-parameters, details are in Appendix B.4. In general, we hope BERT rep-resentations can help encoder understand English sentences better and thus improve translation performance. We also test Multilingual-BERT (Devlin et al., 2019) to see if a multilingual pre-trained model can generalize better to a newly encountered language. The second semi-supervised method we try is again the back-translation method. Sennrich et al. (2016b) has shown that applying this method on NMT obtains larger improvement than applying it on SMT, and it works better than the shallow or deep fusion methods.

---

[9]Because there is no Cherokee BERT. We tried to initialize the decoder embeddings with BERT pre-trained embeddings for Chr-En translation; however, it does not work well.

**Transferring & Multilingual NMT.** Another important line of research is to improve low-resource translation performance by incorporating knowledge from other language pairs. As mentioned in Section 1, Cherokee is the sole member of the southern branch of the Iroquoian language family, so it seems that Cherokee is not "genealogically" related to any high-resource languages in terms of their language family trees. However, it is still interesting to see whether the translation knowledge between other languages and English can help with the translation between Cherokee and English. Hence, in this paper, we will explore two ways of leveraging other language pairs: Transfer learning and Multilingual joint training. Kocmi and Bojar (2018) proposed a simple and effective continual training strategy for the transfer learning of translation models. This method will first train a "parent" model using one language pair until convergence; then continue the training using another language pair, so as to transfer the translation knowledge of the first language pair to the second pair. Johnson et al. (2017) introduced the "many-to-one" and "one-to-many" methods for multilingual joint training of X-En and En-X systems. They achieve this by simply combining training data, except for the "one-to-many" method, every English sentence needs to start with a special token to specify the language to be translated into. We test both the transferring and multilingual methods for Chr-En/En-Chr translations with 4 other X-En/En-X language pairs (X=Czech/German/Russian/Chinese).

## 5 Results

### 5.1 Experimental Details

We randomly sample 5K-100K sentences (about 0.5-10 times the size of the parallel training set) from News Crawl 2017[10] as our English monolingual data. We randomly sample 12K-58K examples (about 1-5 times the size of parallel training set) for each of the 4 language pairs (Czech/German/Russian/Chinese-English) from News Commentary v13 of WMT2018[11] and Bible-uedin (Christodouloupoulos and Steedman, 2015) on OPUS[12]. We apply tokenizer and truecaser from Moses (Koehn et al., 2007). We also apply the BPE tokonization (Sennrich et al., 2016c), but instead of using it as default, we treat it as hyper-parameter. For systems with BERT, we apply the WordPiece tokenizer (Devlin et al., 2019). We compute detokenized and case-sensitive BLEU score (Papineni et al., 2002) using SacreBLEU (Post, 2018).[13]

We implement our SMT systems via Moses (Koehn et al., 2007). **SMT** denotes the base system; **SMT+bigLM** represents the SMT system that uses additional monolingual data to train its language model; SMT with back-translation is denoted by **SMT+BT**. Our NMT systems are implemented by OpenNMT toolkit (Klein et al., 2017). Two baselines are **RNN-NMT** and **Transformer-NMT**. For En-Chr, we also test adding BERT or Multilingual-BERT representations (Devlin et al., 2019), **NMT+BERT** or **NMT+mBERT**, and with back-translation, **NMT+BT**. For Chr-En, we only test **NMT+BT**, treating the English monolingual data size as hyper-parameter. For both En-Chr and Chr-En, we test **T**ransfer learning from and **M**ultilingual joint training with 4 other languages denoted by **NMT+X (T)** and **NMT+X (M)** respectively, where X = Czech/German/Russian/Chinese. We treat the X-En data size as hyper-parameter. All other detailed model designs and hyper-parameters are introduced in Appendix B.

### 5.2 Quantitative Results

Our main experimental results are shown in Table 3 and Table 4.[14] Overall, the translation performance is poor compared with the results of some high-resource translations (Sennrich et al., 2016a), which means that current popular SMT and NMT techniques still struggle to translate well between Cherokee and English especially for the out-of-domain generalization.

---

[10] http://data.statmt.org/news-crawl/en/
[11] http://www.statmt.org/wmt18/index.html
[12] http://opus.nlpl.eu/bible-uedin.php
[13] BLEU+c.mixed+#.1+s.exp+tok.13a+v.1.4.4
[14] The confidence intervals in Table 3 and Table 4 are computed by the bootstrap method (Efron and Tibshirani, 1994).

| ID | System | Cherokee-English | | | | English-Cherokee | | | |
|---|---|---|---|---|---|---|---|---|---|
| | | Dev | Test | Out-dev | Out-test | Dev | Test | Out-dev | Out-test |
| S1 | SMT | 15.0 | 14.5 | 6.7 | 6.4 | 11.1 | 9.8 | 5.4 | 4.7 |
| S2 | + bigLM | 15.3 | 14.5 | **6.8** | **6.5** (±1.4) | 11.3 | 10.1 | 5.4 | 4.7 |
| S3 | + BT | 15.4 | 14.5 | 6.5 | 5.9 | 11.4 | 9.9 | **5.7** | **5.0** (±1.2) |
| N4 | RNN-NMT | 15.7 | 15.1 | 2.7 | 1.8 | 12.4 | 11.7 | 1.1 | 1.8 |
| N5 | + BERT | - | - | - | - | 12.8 | 12.2 | 0.7 | 0.5 |
| N6 | + mBERT | - | - | - | - | 12.4 | 12.0 | 0.5 | 0.4 |
| N7 | + BT | 16.0 | 14.9 | 3.6 | 2.7 | 11.4 | 11.0 | 1.2 | 1.5 |
| N8 | Transformer-NMT | 9.6 | 9.1 | 1.1 | 0.7 | 7.9 | 7.4 | 0.4 | 0.3 |
| N9 | + BERT | - | - | - | - | 8.0 | 7.2 | 0.4 | 0.2 |
| N10 | + mBERT | - | - | - | - | 6.8 | 6.3 | 0.4 | 0.2 |
| N11 | + BT | 9.9 | 9.4 | 1.3 | 0.5 | 6.6 | 5.8 | 0.4 | 0.1 |

Table 3: Performance of our supervised/semi-supervised SMT/NMT systems. **Bold** numbers are our best out-of-domain systems together with Table 4, selected by performance on Out-dev. (±x) shows 95% confidence interval.

**Chr-En vs. En-Chr.** Overall, the Cherokee-English translation gets higher BLEU scores than the English-Cherokee translation. It is reasonable because English has a smaller vocabulary and simpler morphology; thus, it is easier to generate.

**SMT vs. NMT.** For in-domain evaluation, the best NMT systems surpass SMT for both translation directions. It could result from our extensive architecture hyper-parameter search; or, it supports our conjecture that SMT is not necessarily better than NMT because of the different word orders. But, SMT is dominantly better than NMT for out-of-domain evaluation, which is consistent with the results in Koehn and Knowles (2017).

**RNN vs. Transformer.** Transformer-NMT performs worse than RNN-NMT, which contradicts the trends of some high-resource translations (Vaswani et al., 2017). We conjecture that Transformer architecture is more complex than RNN and thus requires larger-scale data to train properly. We also notice that Transformer models are very sensitive to hyper-parameters, so it can be possibly improved after a more extensive hyper-parameter search. The best Transformer-NMT has a 5-layer encoder/decoder and 2-head attention, which is smaller-scale than the model used for high-resource translations (Vaswani et al., 2017). Another interesting observation is that previous works have shown applying BPE and using a small vocabulary by setting minimum word frequency are beneficial for low-resource translation (Sennrich et al., 2016c; Sennrich and Zhang, 2019); however, these techniques are not always being favored during our model selection procedure, as shown in Appendix B.4.

**Supervised vs. Semi-supervised.** As shown in Table 3, using a big language model and back-translation both only slightly improve SMT baselines on both directions. For English-Cherokee translation, leveraging BERT representations improves RNN-NMT by 0.4/0.5 BLEU points on Dev/Test. Multilingual-BERT does not work better than BERT. Back-translation with our Cherokee monolingual data barely improves performance for both in-domain and out-of-domain evaluations, probably because the monolingual data is also out-of-domain, 72% of the unique Cherokee tokens are unseen in the whole parallel data. For Cherokee-English translation, back-translation improves the out-of-domain evaluation of RNN-NMT by 0.9/0.9 BLEU points on Out-dev/Out-test, while it does not obviously improve in-domain evaluation. A possible reason is that the English monolingual data we used is news data that is not of the same domain as Dev/Test but closer to Out-dev/Out-test so that it helps the model to do domain adaptation. We also investigate the influence of the English monolingual data size. We find that all of the NMT+BT systems perform best when only using 5K English monolingual data, see Figure 5 in Appendix B.5.

**Transferring vs. Multilingual.** Table 4 shows the transfer learning and multilingual joint training results. It can be observed that, in most cases, the in-domain RNN-NMT baseline (N4) can be improved by both methods, which demonstrates that even though the 4 languages are not related to Cherokee, their translation knowledge can still be helpful. Transferring from the Chinese-English model and joint training with English-German data achieve our best in-domain Cherokee-English and English-Cherokee performance, respectively. However, there is barely an improvement on the out-of-domain evaluation sets, even though the X-En/En-X data is mostly news (same domain as Out-dev/Out-test). On average, multilingual joint training performs slightly better than transfer learning and usually prefers a larger X-En/En-X data size (see details in Appendix B.4).

| ID | System | Cherokee-English | | | | English-Cherokee | | | |
|---|---|---|---|---|---|---|---|---|---|
| | | Dev | Test | Out-dev | Out-test | Dev | Test | Out-dev | Out-test |
| N4 | RNN-NMT | 15.7 | 15.1 | 2.7 | 1.8 | 12.4 | 11.7 | 1.1 | 1.8 |
| N12 | + Czech (T) | 15.8 | 14.7 | 2.3 | 1.8 | 12.7 | 12,6 | 1.8 | 2.4 |
| N13 | + German (T) | 15.9 | 14.8 | 2.3 | 1.1 | 12.9 | 12.1 | 1.8 | 1.4 |
| N14 | + Russian (T) | 16.5 | 15.8 | 1.9 | 1.9 | 12.6 | 11.8 | 1.8 | 2.3 |
| N15 | + Chinese (T) | **16.9** | **15.8** ($\pm 1.2$) | 2.0 | 1.5 | 12.9 | 12.9 | 1.2 | 0.8 |
| N16 | + Czech (M) | 16.6 | 15.7 | 2.4 | 2.0 | 13.2 | 12.4 | 1.1 | 2.1 |
| N17 | + German (M) | 16.6 | 15.4 | 2.3 | 1.4 | **13.4** | **12.7** ($\pm 1.0$) | 0.8 | 2.0 |
| N18 | + Russian (M) | 16.5 | 15.9 | 1.9 | 1.6 | 13.2 | 13.1 | 1.2 | 1.8 |
| N19 | + Chinese (M) | 16.8 | 16.1 | 2.2 | 1.8 | 13.0 | 13.0 | 1.1 | 1.4 |

Table 4: Performance of our transfer and multilingual learning systems. **Bold** numbers are our best in-domain systems together with Table 3, selected by the performance on Dev. ($\pm x$) shows the 95% confidence interval.

### 5.3 Qualitative Results

Automatic metrics are not always ideal for natural language generation (Wieting et al., 2019). As a new language to the NLP community, we are also not sure if BLEU is a good metric for Cherokee evaluation. Therefore, we conduct a small-scale human (expert) pairwise comparison by our coauthor between the translations generated by our NMT and SMT systems. We randomly sample 50 examples from Test or Out-test, anonymously shuffle the translations from two systems, and ask our coauthor to choose which one they think is better.[15] As shown in Table 5, human preference does not always follow the trends of BLEU scores. For English-Cherokee translation, though the RNN-NMT+BERT (N5) has a better BLEU score than SMT+BT (S3) (12.2 vs. 9.9), it is liked less by humans (21 vs. 29), indicating that BLEU is possibly not a suitable for Cherokee evaluation. A detailed study is beyond the scope of this paper but is an interesting future work direction.

## 6 Conclusion and Future Work

In this paper, we make our effort to revitalize the Cherokee language by introducing a clean Cherokee-English parallel dataset, *ChrEn*, with 14K sentence pairs; and 5K Cherokee monolingual sentences. It not only can be another resource for low-resource machine translation research but also will help to attract attention from the NLP community to save this dying language. Besides, we propose our Chr-En and En-Chr baselines, including both SMT and NMT models, using both supervised and semi-supervised methods, and exploring both transfer learning and multilingual joint training methods with 4 other languages. Experiments show that SMT is significantly better and NMT under out-of-domain condition while NMT is better for in-domain evaluation; and the semi-supervised learning, transfer learning, and multilingual joint training can improve simply supervised baselines. Overall, our best models achieve 15.8/12.7 BLEU for in-domain Chr-En/En-Chr translations and 6.5/5.0 BLEU for out-of-domain Chr-En/En-Chr translations. We hope these diverse baselines will serve as useful strong starting points for future work by the community. Our future work involves converting the monolingual data to parallel and collecting more data from the news domain.

[15]The author, who conducted this human study, was not involved in the development of MT systems.

| Condition | | System IDs | Win | Lose |
|---|---|---|---|---|
| Chr-En | Test | N7 vs. S3 | 43 | 7 |
| | Out-test | N7 vs. S2 | 16 | 34 |
| En-Chr | Test | N5 vs. S3 | 21 | 29 |
| | Out-test | N7 vs. S3 | 2 | 48 |

Table 5: Human comparison between the translations generated from our NMT and SMT systems. If A vs. B, "Win" or "lose" means that the evaluator favors A or B. Systems IDs correspond to the IDs in Table 3.

## Acknowledgments

We thank the reviewers for their helpful feedback and the Kituwah Preservation and Educa-tion Program (KPEP), the Eastern Band of Cherokee Indians, and the Cherokee Nation. This work was supported by NSF-CAREER Award 1846185, ONR Grant N00014-18-1-2871, and faculty awards from Google, Facebook, and Microsoft. The views contained in this article are those of the authors and not of the funding agency.

## References

Elizabeth Albee. 2017. Immersion schools and language learning: A review of cherokee language revitalization efforts among the eastern band of cherokee indians.

Alexei Baevski and Michael Auli. 2019. Adaptive input representations for neural language modeling. In *International Conference on Learning Representations*.

Dzmitry Bahdanau, Kyunghyun Cho, and Yoshua Bengio. 2015. Neural machine translation by jointly learning to align and translate. In *3rd International Conference on Learning Representations, ICLR 2015, San Diego, CA, USA, May 7-9, 2015, Conference Track Proceedings*.

Nicola Bertoldi and Marcello Federico. 2009. Domain adaptation for statistical machine translation with monolingual resources. In *Proceedings of the fourth workshop on statistical machine translation*, pages 182–189.

Jeffrey Bourns. 2019. Cherokee syllabary texts: Digital documentation and linguistic description. In *2nd Conference on Language, Data and Knowledge (LDK 2019)*. Schloss Dagstuhl-Leibniz-Zentrum fuer Informatik.

Christos Christodouloupoulos and Mark Steedman. 2015. A massively parallel corpus: the bible in 100 languages. *Language resources and evaluation*, 49(2):375–395.

David Crystal. 2014. Language Death. Canto Classics. Cambridge University Press.

Ellen Cushman. 2019. Language perseverance and translation of cherokee documents. *College English*, 82(1):115–134.

Jacob Devlin, Ming-Wei Chang, Kenton Lee, and Kristina Toutanova. 2019. Bert: Pre-training of deep bidirectional transformers for language understanding. In *Proceedings of the 2019 Conference of the North American Chapter of the Association for Computational Linguistics: Human Language Technologies, Volume 1 (Long and Short Papers)*, pages 4171–4186.

Kevin Duh, Paul McNamee, Matt Post, and Brian Thompson. 2020. Benchmarking neural and statistical machine translation on low-resource african languages. In *Proceedings of The 12th Language Resources and Evaluation Conference*, pages 2667–2675.

Bradley Efron and Robert J Tibshirani. 1994. *An introduction to the bootstrap*. CRC press.

D. Feeling. 1975. Cherokee-English Dictionary. Cherokee Nation of Oklahoma.

D. Feeling. 1994. A structured approach to learning the basic inflections of the Cherokee verb. Indian University Press, Bacone College.

Durbin Feeling. 2018. *Cherokee Narratives: A Linguistic Study*. University of Oklahoma Press.

Benjamin Frey. 2020. "data is nice:" theoretical and pedagogical implications of an eastern cherokee corpus. *LD&C Special Publication*.

Caglar Gulcehre, Orhan Firat, Kelvin Xu, Kyunghyun Cho, Loic Barrault, Huei-Chi Lin, Fethi Bougares, Holger Schwenk, and Yoshua Bengio. 2015. On using monolingual corpora in neural machine translation. *arXiv preprint arXiv:1503.03535*.

Francisco Guzmán, Peng-Jen Chen, Myle Ott, Juan Pino, Guillaume Lample, Philipp Koehn, Vishrav Chaudhary, and Marc'Aurelio Ranzato. 2019. The flores evaluation datasets for low-resource machine translation: Nepali–english and sinhala–english. In *Proceedings of the 2019 Conference on Empirical Methods in Natural Language Processing and the 9th International Joint Conference on Natural Language Processing (EMNLP-IJCNLP)*, pages 6100–6113.

Kenneth Heafield, Ivan Pouzyrevsky, Jonathan H Clark, and Philipp Koehn. 2013. Scalable modified kneser-ney language model estimation. In *Proceedings of the 51st Annual Meeting of the Association for Computational Linguistics (Volume 2: Short Papers)*, pages 690–696.

Ruth Bradley Holmes and Betty Sharp Smith. 1976. *Beginning Cherokee*. University of Oklahoma Press, Norman.

Sébastien Jean, Orhan Firat, Kyunghyun Cho, Roland Memisevic, and Yoshua Bengio. 2015. Montreal neural machine translation systems for wmt'15. In *Proceedings of the Tenth Workshop on Statistical Machine Translation*, pages 134–140.

Melvin Johnson, Mike Schuster, Quoc V Le, Maxim Krikun, Yonghui Wu, Zhifeng Chen, Nikhil Thorat, Fernanda Viégas, Martin Wattenberg, Greg Corrado, et al. 2017. Google's multilingual neural machine translation system: Enabling zero-shot translation. *Transactions of the Association for Computational Linguistics*, 5:339–351.

M. Joyner. 2014. Cherokee Language Lessons. Lulu Press, Incorporated.

Guillaume Klein, Yoon Kim, Yuntian Deng, Jean Senellart, and Alexander Rush. 2017. OpenNMT: Open-source toolkit for neural machine translation. In *Proceedings of ACL 2017, System Demonstrations*, pages 67–72, Vancouver, Canada. Association for Computational Linguistics.

Tom Kocmi and Ondrej Bojar. 2018. Trivial transfer learning for low-resource neural machine translation. *WMT 2018*, page 244.

Philipp Koehn, Hieu Hoang, Alexandra Birch, Chris Callison-Burch, Marcello Federico, Nicola Bertoldi, Brooke Cowan, Wade Shen, Christine Moran, Richard Zens, et al. 2007. Moses: Open source toolkit for statistical machine translation. In *Proceedings of the 45th annual meeting of the association for computational linguistics companion volume proceedings of the demo and poster sessions*, pages 177–180.

Philipp Koehn and Rebecca Knowles. 2017. Six challenges for neural machine translation. In *Proceedings of the First Workshop on Neural Machine Translation*, pages 28–39.

Philipp Koehn, Franz Josef Och, and Daniel Marcu. 2003. Statistical phrase-based translation. In *Proceedings of the 2003 Conference of the North American Chapter of the Association for Computational Linguistics on Human Language Technology-Volume 1*, pages 48–54. Association for Computational Linguistics.

Surafel M Lakew, Matteo Negri, and Marco Turchi. 2020. Low resource neural machine translation: A benchmark for five african languages. In *AfricaNLP workshop at ICLR 2020*.

Patrik Lambert, Holger Schwenk, Christophe Servan, and Sadaf Abdul-Rauf. 2011. Investigations on translation model adaptation using monolingual data. In *Proceedings of the Sixth Workshop on Statistical Machine Translation*, pages 284–293. Association for Computational Linguistics.

Guillaume Lample, Alexis Conneau, Ludovic Denoyer, and Marc'Aurelio Ranzato. 2018. Unsupervised machine translation using monolingual corpora only. In *International Conference on Learning Representations*.

Minh-Thang Luong, Hieu Pham, and Christopher D Manning. 2015. Effective approaches to attention-based neural machine translation. In *Proceedings of the 2015 Conference on Empirical Methods in Natural Language Processing*, pages 1412–1421.

Brad Montgomery-Anderson. 2008. *A reference grammar of Oklahoma Cherokee*. Ph.D. thesis, University of Kansas.

Cherokee Nation. 2001. Ga-du-gi: A vision for working together to preserve the cherokee language. report of a needs assessment survey and a 10-year language revitalization plan.

Franz Josef Och. 2003. Minimum error rate training in statistical machine translation. In *Proceedings of the 41st Annual Meeting on Association for Computational Linguistics-Volume 1*, pages 160–167. Association for Computational Linguistics.

Franz Josef Och and Hermann Ney. 2003. A systematic comparison of various statistical alignment models. *Computational Linguistics*, 29(1):19–51.

Kishore Papineni, Salim Roukos, Todd Ward, and Wei-Jing Zhu. 2002. Bleu: a method for automatic evaluation of machine translation. In *Proceedings of the 40th annual meeting on association for computational linguistics*, pages 311–318. Association for Computational Linguistics.

M Peake Raymond. 2008. The cherokee nation and its language: Tsalagi ayeli ale uniwonishisdi. *Tahlequah, OK: Cherokee Nation*.

Matt Post. 2018. A call for clarity in reporting bleu scores. In *Proceedings of the Third Conference on Machine Translation: Research Papers*, pages 186–191.

Ofir Press and Lior Wolf. 2017. Using the output embedding to improve language models. In *Proceedings of the 15th Conference of the European Chapter of the Association for Computational Linguistics: Volume 2, Short Papers*, pages 157–163.

Hammam Riza, Michael Purwoadi, Teduh Uliniansyah, Aw Ai Ti, Sharifah Mahani Aljunied, Luong Chi Mai, Vu Tat Thang, Nguyen Phuong Thai, Vichet Chea, Sethserey Sam, et al. 2016. Introduction of the asian language treebank. In *2016 Conference of The Oriental Chapter of International Committee for Coordination and Standardization of Speech Databases and Assessment Techniques (O-COCOSDA)*, pages 1–6. IEEE.

Rico Sennrich, Barry Haddow, and Alexandra Birch. 2016a. Edinburgh neural machine translation systems for wmt 16. In *Proceedings of the First Conference on Machine Translation: Volume 2, Shared Task Papers*, pages 371–376.

Rico Sennrich, Barry Haddow, and Alexandra Birch. 2016b. Improving neural machine translation models with monolingual data. In *Proceedings of the 54th Annual Meeting of the Association for Computational Linguistics (Volume 1: Long Papers)*, pages 86–96.

Rico Sennrich, Barry Haddow, and Alexandra Birch. 2016c. Neural machine translation of rare words with subword units. In *Proceedings of the 54th Annual Meeting of the Association for Computational Linguistics (Volume 1: Long Papers)*, pages 1715–1725.

Rico Sennrich and Biao Zhang. 2019. Revisiting low-resource neural machine translation: A case study. In *Proceedings of the 57th Annual Meeting of the Association for Computational Linguistics*, pages 211–221.

Stephanie Strassel and Jennifer Tracey. 2016. Lorelei language packs: Data, tools, and resources for technology development in low resource languages. In *Proceedings of the Tenth International Conference on Language Resources and Evaluation (LREC'16)*, pages 3273–3280.

Christian Szegedy, Vincent Vanhoucke, Sergey Ioffe, Jon Shlens, and Zbigniew Wojna. 2016. Rethinking the inception architecture for computer vision. In *Proceedings of the IEEE conference on computer vision and pattern recognition*, pages 2818–2826.

Jörg Tiedemann. 2012. Parallel data, tools and interfaces in opus. In *Proceedings of the Eighth International Conference on Language Resources and Evaluation (LREC'12)*, pages 2214–2218.

Hiroto Uchihara. 2016. *Tone and accent in Oklahoma Cherokee*, volume 3. Oxford University Press.

Ashish Vaswani, Noam Shazeer, Niki Parmar, Jakob Uszkoreit, Llion Jones, Aidan N Gomez, Łukasz Kaiser, and Illia Polosukhin. 2017. Attention is all you need. In *Advances in neural information processing systems*, pages 5998–6008.

John Wieting, Taylor Berg-Kirkpatrick, Kevin Gimpel, and Graham Neubig. 2019. Beyond bleu: Training neural machine translation with semantic similarity. In *Proceedings of the 57th Annual Meeting of the Association for Computational Linguistics*, pages 4344–4355.

Shiyue Zhang, Gulnigar Mahmut, Dong Wang, and Askar Hamdulla. 2017. Memory-augmented chinese-uyghur neural machine translation. In *2017 Asia-Pacific Signal and Information Processing Association Annual Summit and Conference (APSIPA ASC)*, pages 1092–1096. IEEE.

Jinhua Zhu, Yingce Xia, Lijun Wu, Di He, Tao Qin, Wengang Zhou, Houqiang Li, and Tieyan Liu. 2020. Incorporating bert into neural machine translation. In *International Conference on Learning Representations*.

# Appendix

## A  Data

### A.1  The Sources of Our Data

Table 14 and Table 15 list the original sources of our parallel and monolingual data, which include the title of original article/book/etc., the name of the speaker/translator, the date of the source, text type, the dialect, and the number of sentences/pairs. OK and NC denote the two existing dialects of Cherokee: the Overhill dialect, most widely spoken in Oklahoma (OK), and the Middle dialect, most widely used in North Carolina (NC).

| Statistics | Ours | OPUS |
|---|---|---|
| Sentence Pairs | 14,101 | 7,974 |
| Dialects | 2 | 1 |
| Text Types | 5 | 2 |
| English tokens | 312,854 | 210,343 |
| Unique English tokens | 13,620 | 6,550 |
| Average English sentence length | 22.2 | 26.4 |
| Cherokee tokens | 205,564 | 144,126 |
| Unique Cherokee tokens | 38,494 | 25,762 |
| Average Cherokee sentence length | 14.6 | 18.1 |

Table 6: The comparison between our parallel data and the data provided on OPUS.

### A.2  Comparison with Existing Data

Here, we compare our parallel data with the data provided on OPUS (Tiedemann, 2012). OPUS has 4 Cherokee-English parallel data resources: Tatoeba, Wikimedia, Bible-uedin, and Ubuntu. Wikimedia's Cherokee sentences are mostly English, and Ubuntu only has several word mappings. We mainly compare with Tatoeba and Bible-uedin. Tatoeba has 22 daily dialogue sentence pairs. Bible-uedin has 15.9K sentence pairs, and after deduplicating,[16] 7.9K pairs are left. It is the translation of the Bible (Cherokee New Testament) (Christodouloupoulos and Steedman, 2015), which is also present in our data. Table 6[17] shows the detailed statistics of our versus OPUS's parallel data. In summary, 99% of the OPUS data is also present in our parallel data, i.e., our data has 6K more sentence pairs that are not sacred texts (novels, news, etc.).

## B  Experimental Details

### B.1  Data and Preprocessing

For semi-supervised learning, we sample additional English monolingual data from News Crawl 2017.[18] We randomly sample 5K, 10K, 20K, 50K, and 100K sentences, which are about half, equal, double, 5-times, 10-times the size of the parallel training set. For transfer and multilingual training experiments, we use 12K, 23K, or 58K X-En (X=Czech/German/Russian/Chinese) parallel examples, which are equal, double, and 5-times the size of Chr-En training set. We sample these examples either only from News Commentary v13 of WMT2018[19] or from both News Commentary and Bible-uedin (Christodouloupoulos and Steedman, 2015) on OPUS[20], because half of in-domain Chr-En data is the Bible. Whenever we sample from Bible-uedin, we keep the sample size as 6K and sample the rest from News Commentary.

---

[16] The deduplication is based on Cherokee sentences not sentence pairs, because we notice that in Bible-uedin two English sentences can be different just because of one additional white space in the sentence.

[17] Note that we apply the same deduplication on our data, so the numbers are slightly different from those in Table 2.

[18] http://data.statmt.org/news-crawl/en/

For all the data we used, the same tokenizer and truecaser from Moses (Koehn et al., 2007) are applied. For some NMT systems, we also apply the BPE subword tokenization (Sennrich et al., 2016c) with 20,000 merge operations for Cherokee and English separately. For NMT systems with BERT, we apply the WordPiece tokenizer from BERT (Devlin et al., 2019) for English. Before evaluation, the translation outputs are detokenized and detruecased. We use SacreBLEU (Post, 2018)[21] to compute the BLEU (Papineni et al., 2002) scores of all translation systems.

### B.2 SMT Systems

We implement SMT systems via Moses (Koehn et al., 2007). We train a 3-gram langauge model (LM) by KenLM (Heafield et al., 2013) and conduct word alignment by GIZA++ (Och and Ney, 2003). Model weights are tuned on the Dev or Our-dev by MERT (Och, 2003).

### B.3 NMT Systems

Our NMT systems are all implemented by Open-NMT (Klein et al., 2017). As shown in Table 3 and Table 4, there are 16 NMT systems in total (N4-N19). For each of these systems, We conduct a limited amount of hyper-parameter grid search on Dev or Out-dev. The search space includes applying BPE or not, minimum word frequency threshold, number of encoder/decoder layers, hidden size, dropout, etc. The detailed hyper-parameter tuning procedure will be discussed in the next subsection. During decoding, all systems use beam search with beam size 5 and replace unknown words with source words that have the highest attention weight.

### B.4 Hyper-parameters

We observed the NMT models, especially the Transformer-NMT models, are sensitive to hyper-parameters. Thus, we did a limited amount of hyper-parameter grid search when developing NMT models. For building vocabulary, we take BPE (Sennrich et al., 2016c) (use or not) and the minimum word frequency (0, 5, 10) as two hyper-parameters. For the model architecture, we explore different number of encoder/decoder layers (1, 2, 3 for RNN; 4, 5, 6 for Transformer), hidden size (512, 1024), embedding size (equals to hidden size, except 768 for BERT), tied decoder embeddings (Press and Wolf, 2017) (use or not), and number of attention heads (2, 4, 8). For training techniques, we tune dropout (0.1, 0.2, 0.3), label smoothing (Szegedy et al., 2016) (0.1, 0.2), average decay (1e-4 or not use), batch type (tokens or sentences), batch size (1000, 4000 for tokens; 32, 64 for sents), and warmup steps (3000, 4000). We take the English monolingual data size (5K, 10K, 20K, 50K, 100K) as hyper-parameter when we do back-translation for Cherokee-English translation. We take the size of Czech/German/Russian/Chinese-English parallel data (12K, 23K, 58K) and whether sampling from Bible-uedin (yes or no) as hyper-parameter when we do transfer or multilingual training. Besides, we take how we incorporate BERT as hyper-parameter, and it is chosen from the following five settings and their combinations:

- BERT embedding: Initializing NMT models' word embedding matrix with BERT's pretrained word embedding matrix $I_B$, corresponding to ① in Figure 4;

- BERT embedding (fix): The same as "BERT embedding" except we fix the word embedding during training;

- BERT input: Concatenate NMT encoder's input $I_E$ with BERT's output $H_B$, corresponding to ② in Figure 4;

- BERT output: Concatenate NMT encoder's output $H_E$ with BERT's output $H_B$, corresponding to ③ in Figure 4;

- BERT output (attention): Use another attention to leverage BERT's output $H_B$ into decoder, corresponding to ④ in Figure 4;

"BERT embedding" and "BERT embedding (fix)" will not be applied simultaneously, and "BERT output" and "BERT output (attention)" will not be applied simultaneously. Multilingual-BERT is used in the same ways. At most, there

---

[19] http://www.statmt.org/wmt18/index.html
[20] http://opus.nlpl.eu/bible-uedin.php
[21] BLEU+c.mixed+#.1+s.exp+tok.13a+v.1.4.4

|  | Dev | | | | Out-dev | | | |
| --- | --- | --- | --- | --- | --- | --- | --- | --- |
| Hyper-parameter | N4 | N7 | N8 | N11 | N4 | N7 | N8 | N11 |
| BPE | yes | | | | - | | | |
| word min frequency | 10 | | | | 0 | | 10 | |
| encoder layer | 2 | | 5 | | 2 | | 5 | |
| decoder layer | 2 | | 5 | | 2 | | 5 | |
| hidden size | 1024 | | | | 512 | | 1024 | |
| embedding size | 1024 | | | | 512 | | 1024 | |
| tied decoder embeddings | yes | | - | | yes | | - | yes |
| head | - | | 2 | | - | | 2 | |
| dropout | 0.3 | 0.5 | 0.1 | | 0.3 | | 0.1 | |
| label smoothing | 0.2 | | 0.1 | | 0.2 | | 0.1 | |
| average decay | 1e-4 | | - | | 1e-4 | | | |
| batch type | tokens | sents | tokens | | sents | | tokens | |
| batch size | 1000 | 32 | 4000 | | 32 | | 4000 | |
| optimizer | adam | | | | | | | |
| learning rate (lr) | 5e-4 | | | | | | | |
| lr decay method | - | | rsqrt | | - | | rsqrt | |
| warmup steps | - | | 4000 | | - | | 4000 | |
| early stopping | 10 | | | | | | | |
| mono. data size | - | 5000 | - | 5000 | - | 5000 | - | 5000 |

Table 7: The hyper-parameter settings of **Supervised and Semi-supervised Cherokee-English NMT systems in Table 3**. Empty fields indicate that hyper-parameter is the same as the previous (left) system.

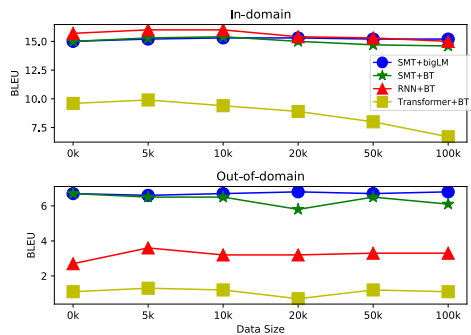

Figure 5: The influence of the English monolingual data size on semi-supervised learning performance. The results are on Dev or Out-dev.

### B.5 English Monolingual Data Size Influence

In the semi-supervised experiments of Cherokee-English, we investigate the influence of the English monolingual data size. As mentioned above, we use 5K, 10K, 20K, 50K, and 100K English monolingual sentences. Figure 5 shows its influence on translation performance. It can be observed that increasing English monolingual data size does not lead to higher performance, especially, all NMT+BT systems achieve the best performance when only use 5K English sentences.

are 576 searches per model, but oftentimes, we did less than that because we early cut off unpromising settings. All hyper-parameters are tuned on Dev or Out-dev for in-domain or out-of-domain evaluation, and the model selection is based on translation accuracy on Dev or Out-dev. Table 7, Table 8, Table 9, Table 10, Table 11, Table 12, and Table 13 list the hyper-parameters of all the systems shown in the Table 3 and Table 4. Since our parallel dataset is small (14K sentence pairs), even the slowest experiment, Transformer-NMT+mBERT, only takes 2 minutes per epoch using one Tesla V100 GPU. We train 100 epochs at most and using early stop when the translation accuracy on Dev or Out-dev does not improve for 10 epochs.

|  | Dev | | | | Out-dev | | | |
| --- | --- | --- | --- | --- | --- | --- | --- | --- |
| Hyper-parameter | N12 | N13 | N14 | N15 | N12 | N13 | N14 | N15 |
| BPE | - | | | | | | | |
| word min frequency | 0 | | | | | 5 | | |
| encoder layer | 2 | | | | | | | |
| decoder layer | 2 | | | | | | | |
| hidden size | 1024 | | | | 512 | | | |
| embedding size | 1024 | | | | 512 | | | |
| tied decoder embeddings | yes | | | | | | | |
| head | | | | | | | | |
| dropout | 0.3 | | | | | | | |
| label smoothing | 0.2 | | | | | | | |
| average decay | 1e-4 | | | | | | | |
| batch type | tokens | | | | sents | | | |
| batch size | 1000 | | | | 32 | | | |
| optimizer | adam | | | | | | | |
| learning rate (lr) | 5e-4 | | | | | | | |
| lr decay method | - | | | | | | | |
| warmup steps | - | | | | | | | |
| early stopping | 10 | | | | | | | |
| X-En data size | 11,639 | | | 23,278 | 11,639 | | | 23,278 |
| with Bible | no | | yes | | no | | yes | |

Table 8: The hyper-parameter settings of **Transferring Cherokee-English NMT systems in Table 4**. Empty fields indicate that hyper-parameter is the same as the previous (left) system.

|  | Dev | | | | Out-dev | | | |
| --- | --- | --- | --- | --- | --- | --- | --- | --- |
| Hyper-parameter | N16 | N17 | N18 | N19 | N16 | N17 | N18 | N19 |
| BPE | - | | | | | | | |
| word min frequency | 5 | | | | | 5 | | |
| encoder layer | 2 | | | | | | | |
| decoder layer | 2 | | | | | | | |
| hidden size | 1024 | | | | 512 | | | |
| embedding size | 1024 | | | | 512 | | | |
| tied decoder embeddings | yes | | | | | | | |
| head | | | | | | | | |
| dropout | 0.3 | | | | | | | |
| label smoothing | 0.2 | | | | | | | |
| average decay | 1e-4 | | | | | | | |
| batch type | tokens | | | | sents | | | |
| batch size | 1000 | | | | 32 | | | |
| optimizer | adam | | | | | | | |
| learning rate (lr) | 5e-4 | | | | | | | |
| lr decay method | - | | | | | | | |
| warmup steps | - | | | | | | | |
| early stopping | 10 | | | | | | | |
| X-En data size | 58,195 | | | | 23,278 | | | |
| with Bible | yes | | | | no | | | |

Table 9: The hyper-parameter settings of **Multilingual Cherokee-English NMT systems in Table 4**. Empty fields indicate that hyper-parameter is the same as the previous (left) system.

|  | Dev | | | | | | | |
| --- | --- | --- | --- | --- | --- | --- | --- | --- |
| Hyper-parameter | N4 | N7 | N5 | N6 | N8 | N11 | N9 | N10 |
| BPE | - |  | - |  | yes |  | - |  |
| WordPiece | - |  | yes |  | - |  | yes |  |
| word min frequency | 0 |  |  |  | 5 |  | 0 |  |
| encoder layer | 2 |  |  |  | 5 |  |  |  |
| decoder layer | 2 |  |  |  | 5 |  |  |  |
| hidden size | 1024 |  |  |  |  |  |  |  |
| embedding size | 1024 |  | 768 |  | 1024 |  | 768 |  |
| tied decoder embeddings | yes |  |  |  | - |  |  |  |
| head | - |  |  |  | 2 |  |  |  |
| dropout | 0.5 |  |  |  | 0.1 |  |  |  |
| label smoothing | 0.2 |  |  |  | 0.1 | 0.2 | 0.1 |  |
| average decay | 1e-4 |  |  |  | - |  |  |  |
| batch type | tokens |  |  |  |  |  |  |  |
| batch size | 1000 |  |  |  | 4000 |  |  |  |
| optimizer | adam |  |  |  |  |  |  |  |
| learning rate (lr) | 5e-4 |  |  |  |  |  |  |  |
| lr decay method | - |  |  |  | rsqrt |  |  |  |
| warmup steps | - |  |  |  | 4000 |  |  |  |
| early stopping | 10 |  |  |  |  |  |  |  |
| mono. data size | - | 5210 | - |  |  | 5210 | - |  |
| BERT embedding | - |  |  |  |  |  | yes |  |
| BERT embedding (fix) | - |  | yes | - |  | - |  |  |
| BERT input | - |  | yes |  | - |  | yes |  |
| BERT output | - |  | yes | - |  |  | yes |  |
| BERT output (attention) | - |  |  |  | - |  |  |  |

Table 10: The hyper-parameter settings of **in-domain Supervised and Semi-supervised English-Cherokee NMT systems in Table 3**. Empty fields indicate that hyper-parameter is the same as the previous (left) system.

|  | Out-dev | | | | | | | |
| --- | --- | --- | --- | --- | --- | --- | --- | --- |
| Hyper-parameter | N4 | N7 | N5 | N6 | N8 | N11 | N9 | N10 |
| BPE | - |  |  |  |  |  |  |  |
| WordPiece | - |  | yes |  | - |  | yes |  |
| word min frequency | 10 |  |  | 0 | 0 |  |  |  |
| encoder layer | 2 |  |  |  | 5 |  |  |  |
| decoder layer | 2 |  |  |  | 5 |  |  |  |
| hidden size | 512 |  |  |  | 1024 |  |  |  |
| embedding size | 512 |  | 768 |  | 1024 |  | 768 |  |
| tied decoder embeddings | yes |  |  |  | - | yes | - |  |
| head | - |  |  |  | 2 |  |  |  |
| dropout | 0.3 | 0.5 | 0.3 |  | 0.1 |  |  |  |
| label smoothing | 0.2 | 0.1 | 0.2 |  | 0.2 |  |  |  |
| average decay | 1e-4 |  |  |  | - | 1e-4 | - |  |
| batch type | sents |  |  |  | tokens |  |  |  |
| batch size | 32 |  |  |  | 4000 |  |  |  |
| optimizer | adam |  |  |  |  |  |  |  |
| learning rate (lr) | 5e-4 |  |  |  |  |  |  |  |
| lr decay method | - |  |  |  | rsqrt |  |  |  |
| warmup steps | - |  |  |  | 4000 |  |  |  |
| early stopping | 10 |  |  |  |  |  |  |  |
| mono. data size | - | 5210 | - |  |  | 5210 | - |  |
| BERT embedding | - |  | yes |  | - |  |  |  |
| BERT embedding (fix) | - |  |  |  |  |  | yes | - |
| BERT input | - |  | yes |  | - |  |  |  |
| BERT output | - |  |  |  |  |  | yes | - |
| BERT output (attention) | - |  |  |  |  |  |  |  |

Table 11: The hyper-parameter settings of **out-of-domain Supervised and Semi-supervised English-Cherokee NMT systems in Table 3**. Empty fields indicate that hyper-parameter is the same as previous (left) system.

|  | Dev | | | | Out-dev | | | |
| --- | --- | --- | --- | --- | --- | --- | --- | --- |
| Hyper-parameter | N12 | N13 | N14 | N15 | N12 | N13 | N14 | N15 |
| BPE | - | | | | | | | |
| word min frequency | 0 | | | | 10 | 5 | 10 | |
| encoder layer | 2 | | | | | | | |
| decoder layer | 2 | | | | | | | |
| hidden size | 1024 | | | | 512 | | | |
| embedding size | 1024 | | | | 512 | | | |
| tied decoder embeddings | yes | | | | | | | |
| head | | | | | | | | |
| dropout | 0.3 | | | | | | | |
| label smoothing | 0.2 | | | | | | | |
| average decay | 1e-4 | | | | | | | |
| batch type | tokens | | | | sents | | | |
| batch size | 1000 | | | | 32 | | | |
| optimizer | adam | | | | | | | |
| learning rate (lr) | 5e-4 | | | | | | | |
| lr decay method | - | | | | | | | |
| warmup steps | - | | | | | | | |
| early stopping | 10 | | | | | | | |
| En-X data size | 23,278 | 11,639 | 23,278 | | 11,639 | 23,278 | | 11,639 |
| with Bible | yes | no | yes | | | no | | |

Table 12: The hyper-parameter settings of **Transferring English-Cherokee NMT systems in Table 4**. Empty fields indicate that hyper-parameter is the same as the previous (left) system.

|  | Dev | | | | Out-dev | | | |
| --- | --- | --- | --- | --- | --- | --- | --- | --- |
| Hyper-parameter | N16 | N17 | N18 | N19 | N16 | N17 | N18 | N19 |
| BPE | - | | | | | | | |
| word min frequency | 5 | | | | 5 | | | |
| encoder layer | 2 | | | | | | | |
| decoder layer | 2 | | | | | | | |
| hidden size | 1024 | | | | 512 | | | |
| embedding size | 1024 | | | | 512 | | | |
| tied decoder embeddings | yes | | | | | | | |
| head | | | | | | | | |
| dropout | 0.3 | | | | | | | |
| label smoothing | 0.2 | | | | | | | |
| average decay | 1e-4 | | | | | | | |
| batch type | tokens | | | | sents | | | |
| batch size | 1000 | | | | 32 | | | |
| optimizer | adam | | | | | | | |
| learning rate (lr) | 5e-4 | | | | | | | |
| lr decay method | - | | | | | | | |
| warmup steps | - | | | | | | | |
| early stopping | 10 | | | | | | | |
| En-X data size | 58,195 | | | | 23,278 | | | 11,639 |
| with Bible | yes | | | | no | | yes | no |

Table 13: The hyper-parameter settings of **Multilingual English-Cherokee NMT systems in Table 4**. Empty fields indicate that hyper-parameter is the same as the previous (left) system.

| Title | Speaker/Translator | Date | Type | Dialect | Examples |
| --- | --- | --- | --- | --- | --- |
| Cherokee New Testament | Elias Boudinot & Samuel Worcester | 1860 | Sacred Text | OK | 7,957 |
| Charlotte's Web | Myrtle Driver Johnson | 2015 | Novel | NC | 3,029 |
| Thirteen Moons | Myrtle Driver Johnson | 2007 | Novel | NC | 1,927 |
| A Walk in the Woods | Marie Junaluska | 2011 | Children's nonfiction | NC | 104 |
| Wolf Wears Shoes (from: Cherokee Stories of the Turtle Island Liars' Club) | Sequoyah Guess | 2012 | Traditional narrative | OK | 97 |
| The Big Journey of Little Fish | Myrtle Driver Johnson & Abel Catolster | 2010 | Children's fiction | NC | 97 |
| NSU to host 2017 Inter-Tribal Language Summit | David Crawler | 2017 | News article | OK | 69 |
| Bobby the Bluebird - The Blizzard Blunder | Myrtle Driver Johnson | 2016 | Children's fiction | NC | 66 |
| A Very Windy Day | Myrtle Driver Johnson | 2011 | Children's fiction | NC | 59 |
| Sequoyah: The Cherokee Man Who Gave His People Writing | Anna Sixkiller Huckaby | 2004 | Children's nonfiction | OK | 56 |
| Spearfinger | Nannie Taylor | 2008 | Traditional narrative | NC | 50 |
| Tom Belt Meets Horse | Tom Belt | 2008 | Personal Narrative | OK | 45 |
| The Beast | Marie Junaluska | 2012 | Children's fiction | NC | 45 |
| Jackson waiting for lung, heart transplants | Anna Sixkiller Huckaby | 2017 | News article | OK | 42 |
| Hannah creates competitive softball league | Anna Sixkiller Huckaby | 2017 | News article | OK | 39 |
| CN re-opens Sequoyah's Cabin Museum | Anna Sixkiller Huckaby | 2017 | News article | OK | 36 |
| Chance finds passion in creating soap | Anna Sixkiller Huckaby | 2016 | News article | OK | 36 |
| Ice passes on loom weaving knowledge | David Crawler | 2017 | News article | OK | 35 |
| Cherokee National Holiday sees first-ever chunkey game | Anna Sixkiller Huckaby | 2017 | News article | OK | 34 |
| Gonzales showcases interpretive Cherokee art | David Crawler | 2017 | News article | OK | 33 |
| Eating healthy on a budget | David Crawler | 2017 | News article | OK | 31 |
| Team Josiah fundraises for diabetes awareness | Anna Sixkiller Huckaby | 2017 | News article | OK | 30 |
| Cherokee Gates scholars reflect on program's influence | Anna Sixkiller Huckaby | 2017 | News article | OK | 28 |
| 'Mankiller' premieres June 19 at LA Film Festival | Anna Sixkiller Huckaby | 2017 | News article | OK | 26 |
| Hummingbird, Dart named Cherokee National Treasures | Dennis Sixkiller | 2017 | News article | OK | 25 |
| CNF scholarship applications open Nov. 1 | Anna Sixkiller Huckaby | 2017 | News article | OK | 22 |
| Chunestudy feels at home as CHC curator | Anna Sixkiller | 2016 | News article | OK | 20 |
| One Time in Chapel Hill… | Tom Belt | 2008 | Personal Narrative | OK | 20 |
| Ball of Fire (From: Cherokee Narratives: A Linguistic Study) | Durbin Feeling | 2018 | Personal Narrative | OK | 20 |
| Cat Meowing (From: Cherokee Narratives: A Linguistic Study) | Durbin Feeling | 2018 | Personal Narrative | OK | 19 |
| Peas – Our Garden, Our Life | Marie Junaluska | 2013 | Children's nonfiction | NC | 18 |
| Stopping by Woods on a Snowy Evening | Marie Junaluska | 2017 | Poetry | NC | 16 |
| The Invisible Companion Fox (From: Cherokee Narratives: A Linguistic Study) | Durbin Feeling | 2018 | Personal Narrative | OK | 14 |
| Cherokee Speakers Bureau set for April 12 | Anna Sixkiller Huckaby | 2018 | News article | OK | 6 |

Table 14: Parallel Data Sources.

| Title | Speaker/Translator | Date | Type | Dialect | Examples |
|---|---|---|---|---|---|
| Cherokee Old Testament | Samuel Worcester | 1860 | Sacred Text | OK | 3802 |
| Encyclopedia Brown | Marie Junaluska | 2016 | Novel | NC | 537 |
| Charlie Brown Christmas | Wiggins Blackfox | 2020 | Children's fiction | NC | 146 |
| Interview with Wilbur Sequoyah | Durbin Feeling | 2018 | Dialogue | OK | 96 |
| One Fish Two Fish Red Fish Blue Fish | Marie Junaluska | 2019 | Children's Fiction | NC | 91 |
| Climbing The Apple Tree | Marie Junaluska | 2020 | Children's Nonfiction | NC | 59 |
| How Jack Went to Seek His Fortune | Wiggins Blackfox | 2019 | Children's Fiction | NC | 50 |
| Trick Or Treat Danny | Wiggins Blackfox | 2019 | Children's Fiction | NC | 49 |
| Kathy's Change | Myrtle Driver Johnson | 2016 | Children's Fiction | NC | 45 |
| Crane And Hummingbird Race | Dennis Sixkiller | 2007 | Traditional Narrative | OK | 44 |
| Ten Apples On Top | Myrtle Driver Johnson | 2017 | Children's Fiction | NC | 37 |
| Transformation | Durbin Feeling | 2018 | Personal Narrative | OK | 35 |
| Halloween | Wiggins Blackfox | 2019 | Children's Fiction | NC | 26 |
| Throw It Home | Mose Killer | 2018 | Personal Narrative | OK | 21 |
| Little People | Durbin Feeling | 2018 | Personal Narrative | OK | 19 |
| Hunting Dialogue | Durbin Feeling | 2018 | Dialogue | OK | 18 |
| Two Dogs in On | Durbin Feeling | 2018 | Personal Narrative | OK | 18 |
| Reminiscence | Mose Killer | 2018 | Personal Narrative | OK | 17 |
| The Origin of Evil Magic | Homer Snell | 2018 | Personal Narrative | OK | 17 |
| Water Beast | Sam Hair | 2018 | Personal Narrative | OK | 16 |
| Legal Document | John Littlebones | 2018 | Personal Narrative | OK | 14 |
| The Good Samaritan | Samuel Worcester | 1860 | Sacred Text | OK | 12 |
| My Grandma | Wiggins Blackfox | 2018 | Children's Nonfiction | NC | 9 |
| Rabbit and Buzzard | Charley Campbell | 2018 | Personal Narrative | OK | 7 |
| Hello Beach | Marie Junaluska | 2020 | Children's Nonfiction | NC | 7 |
| This Is My Little Brother | Marie Junaluska | 2017 | Children's Fiction | NC | 7 |
| Diary | Author Unknown | 2018 | Personal Narrative | OK | 6 |
| How to Make Chestnut Bread | Annie Jessan | 2018 | Personal Narrative | OK | 5 |

Table 15: Monolingual Data Sources.